\documentclass[10pt,journal]{IEEEtran}
\makeatletter
\usepackage{amsmath,amsfonts}
\usepackage{algorithm}
\usepackage{array}
\usepackage{textcomp}
\usepackage{stfloats}
\usepackage{url}
\usepackage{verbatim}
\usepackage[table]{xcolor}

\definecolor{verylightgray}{gray}{0.9} 

\usepackage{graphicx}
\usepackage{cite}
\usepackage{booktabs}
\usepackage{multirow}
\usepackage{bbding}
\usepackage{subcaption}
\usepackage{algorithm}
\usepackage{algpseudocode}
\usepackage{amsmath} 
\usepackage{amssymb} 
\usepackage{tabularx, booktabs}
\usepackage[pagebackref=true,breaklinks=true,colorlinks,citecolor=blue,linkcolor=blue,bookmarks=false]{hyperref}
\usepackage[table]{xcolor}
\begin{document}

\title{Entity-Guided Multi-Task Learning for Infrared and Visible Image Fusion}

\author{Wenyu Shao, Hongbo Liu, Yunchuan Ma, Ruili Wang




}

\markboth{IEEE Transactions on Multimedia}%
{Shell \MakeLowercase{\textit{et al.}}: A Sample Article Using IEEEtran.cls for IEEE Journals}

\maketitle

\begin{abstract}

Existing text-driven infrared and visible image fusion approaches often rely on textual information at the sentence level, which can lead to semantic noise from redundant text and fail to fully exploit the deeper semantic value of textual information. To address these issues, we propose a novel fusion approach named Entity-Guided Multi-Task learning for infrared and visible image fusion (EGMT). Our approach includes three key innovative components: (i) A principled method is proposed to extract entity-level textual information from image captions generated by large vision-language models, eliminating semantic noise from raw text while preserving critical semantic information; (ii) A parallel multi-task learning architecture is constructed, which integrates image fusion with a multi-label classification task. By using entities as pseudo-labels, the multi-label classification task provides semantic supervision, enabling the model to achieve a deeper understanding of image content and significantly improving the quality and semantic density of the fused image; (iii) An entity-guided cross-modal interactive module is also developed to facilitate the fine-grained interaction between visual and entity-level textual features, which enhances feature representation by capturing cross-modal dependencies at both inter-visual and visual-entity levels. To promote the wide application of the entity-guided image fusion framework, we release the entity-annotated version of four public datasets (\textit{i.e.}, TNO, RoadScene, M\textsuperscript{3}FD, and MSRS). Extensive experiments demonstrate that EGMT achieves superior performance in preserving salient targets, texture details, and semantic consistency, compared to the state-of-the-art methods. The code and dataset will be publicly available at \url{https://github.com/wyshao-01/EGMT}.

\end{abstract}


\section{Introduction}

\IEEEPARstart{I}nfrared and visible image fusion aims to combine the strengths of each modality to generate an enhanced image, which is crucial for various applications, such as traffic monitoring\cite{CEFusion}, environmental surveillance\cite{environment}, and autonomous driving\cite{autodriven}. Infrared images are capable of capturing the thermal radiation emitted by objects, providing significant heat source perception even in low light or nighttime conditions, while visible images supply abundant color and texture information. By fusing the information of the two modalities, the shortcomings of a single modality can be made up and the perception ability of complex scenes can be improved.

\begin{figure*}[ht]
  \centering
  \includegraphics[width=\linewidth]{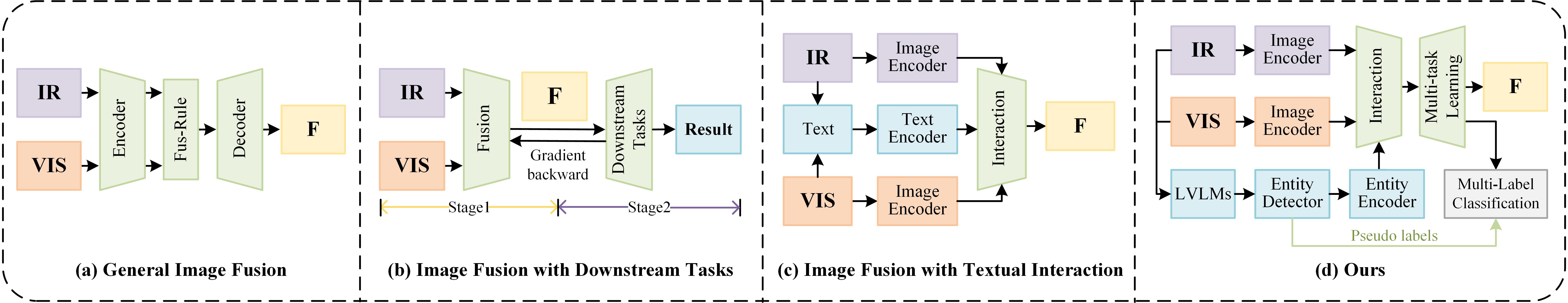}
  \caption{Illustration of (a) General image fusion approach, 
          (b) Cascaded downstream-task image fusion approach, and 
          (c) Textual interaction-based image fusion approach. 
          (d) Our EGMT framework jointly leverages entity semantics 
          and visual features to simultaneously optimize image fusion 
          (main task) and multi-label classification (auxiliary task) 
          through multi-task learning.}
  \label{fig:insight}
\end{figure*}

With the development of neural networks and computational hardware, deep learning-based image fusion approaches have become widely adopted\cite{review}. Compared to traditional approaches\cite{Bayesian,mdlatlrr,tradition1,ICAF}, deep learning-based approaches can automatically learn more complex features from images. General image fusion approaches\cite{dense,DATFuse,swin,AITFuse} commonly adopt an encoder-decoder architecture, where the encoder extracts visual features from both infrared and visible modalities, and the decoder generates the fused image by merging these visual features, as illustrated in Fig.~\ref{fig:insight}~(a). DATFuse~\cite{DATFuse} employs a dual-attention residual module to extract salient features and utilizes a transformer module to supplement global information. Although these methods are effective at extracting visual features, they often neglect high-level semantic information, which can lead to a lack of interpretability of fused results and limit the ability to capture contextual information essential for downstream tasks.

Several works \cite{SEA,TAR,meta,spdfusion,hitfusion} include semantic information into fusion by cascading high-level vision tasks, as illustrated in Fig.~\ref{fig:insight}~(b). A typical example, MetaFusion\cite{meta} generates semantic features compatible with the fusion network by simulating meta-learning, achieving mutually promoting learning between image fusion and object detection tasks. Another example, UAAFusion\cite{UAA} leverages attribution analysis to steer image fusion toward semantically discriminative features, enhancing downstream segmentation performance. These approaches have demonstrated promising results by leveraging semantic information from high-level vision tasks, but they also come with certain limitations. Specifically, while these approaches effectively integrate semantic information through task cascading, they often focus on visual semantics derived from specific tasks. Besides, these approaches still overlook the deeper textual semantic information, and the design of cascading with specific tasks may cause the learned features to be less adaptable to other types of downstream tasks.

Recently, some studies make efforts to leverage textual information to enhance the performance of image fusion. As shown in Fig.~\ref{fig:insight}~(c), this approach utilizes large vision-language models (LVLMs) or manual annotation to generate image captions or prompts-based sentences, and then leverage the generated textual information to guide the image fusion. For example, FILM~\cite{FILM} first uses several pre-trained multi-modal models to generate paragraph-level textual information to guide and enhance image fusion, achieving a more comprehensive understanding of image content. Although these approaches leverage deeper textual information beyond vision to enhance fusion results, they still face two major challenges: (i) During the interaction between textual and visual features, certain semantically redundant elements (\textit{e.g.}, article ``the" and preposition ``of") can introduce semantic noise\cite{concept}, thereby reducing the fusion performance. These elements lack contextual relevance in the image domain and may disrupt the fine-grained details of the fusion process, leading to inaccurate semantic representations or noisy artifacts in the final fused outputs. (ii) Simple feature interactions fail to fully leverage the potential of textual information. Traditional networks are constrained solely by pixel-level fusion loss functions, which neglect text-level semantic supervision. These challenges result in suboptimal exploitation of textual information, as the fusion process does not adequately incorporate higher-level contextual constraints.

To address these challenges, we propose a pioneering \textbf{E}ntity-\textbf{G}uided \textbf{M}ulti-\textbf{T}ask (EGMT) learning framework, which leverages entity information to strengthen original visual features and enhance the model's awareness of the entities. Specifically, to address the challenge (i), we build an entity extraction pipeline that uses a pre-trained entity detection model to extract key entities from LVLM-generated captions, eliminating semantically redundant elements of the sentence-level text. For the challenge (ii), we utilize entities as pseudo-labels to construct the multi-label classification as an auxiliary task that boosts the model's understanding of entities. Additionally, we design an entity-guided cross-modal interaction module to facilitate fine-grained entity-visual feature interaction. As depicted in Fig.~\ref{fig:insight}~(d), we first extract entities from the captions of source images using an entity detector, then utilize the encoded entity features to interact with visual features, and finally employ these entities as pseudo-labels for multi-task learning. More specific details of the proposed EGMT are in Sec.~\ref{sec:method}.

Our main contributions are summarized as follows:

\begin{itemize}
\item  We propose a novel entity-guided multi-task learning framework that pioneers entity-level semantic guidance through: (i) Semantic purification via entity extraction from LVLM-generated captions to eliminate textual semantic noise; (ii) An entity-guided cross-modal interaction module enabling fine-grained interaction of visual features and entity semantics.

\item We develop a parallel multi-task architecture that synergistically combines image fusion with multi-label classification. By utilizing extracted entities as pseudo-labels for semantic supervision, the framework achieves dual optimization of pixel-level reconstruction and entity-level semantic understanding.

\item We construct a comprehensive visual-entity benchmark for infrared and visible image fusion, integrating entity annotations with four public datasets (\textit{i.e.}, TNO, RoadScene, M\textsuperscript{3}FD, and MSRS). This resource enables entity-aware fusion research and facilitates future exploration of semantic-guided fusion paradigms.
\end{itemize}

The remainder structure of this paper is as follows: Sec.~\ref{sec:related} provides an overview of the advancements in deep learning-based image fusion approaches, multi-task learning vision tasks, large vision-language models, and entity extraction techniques. Sec.~\ref{sec:method} details the overall architecture of EGMT. Sec.~\ref{sec:experiment} and Sec.~\ref{sec:conclusion} showcase experimental analyses and conclude with pertinent findings, respectively.

\section{Related Work}~\label{sec:related}
\subsection{Text-driven image fusion methods}

The text-driven image fusion methods~\cite{FILM,TEXTFUSION,textif, Text-diffuse,lure, TFD2Fusion} integrate high-level semantic information from textual descriptions to achieve controllability in the fusion process. TextFusion~\cite{TEXTFUSION} embeds textual semantics into the image fusion workflow, thus advancing the control and evaluation of the resulting fused images. Text-IF \cite{textif} employs sentence-level semantic guidance to perceive degradation and achieve interactive image fusion. Text-DiFuse~\cite{Text-diffuse} enables user-controllable foreground enhancement through text-driven zero-shot localization. It generates precise masks and injects a contrast prior to re-modulate foreground regions during diffusion fusion, thereby dynamically amplifying target saliency and semantic fidelity. LURE\cite{lure} reassociates visual features into a unified latent feature space and uses text-guided attention to enhance degradation-aware fusion and downstream task performance. $\text{TFD}^2\text{Fusion}$~\cite{TFD2Fusion} decomposes source images into multi-grained frequency components via customized low- and high-pass masks, then enriches the fused representation by cross-modal interaction with scene- and detail-level textual priors, achieving superior fusion quality under joint spatial- and frequency-domain supervision. 

The above methods utilize crude textual information to exploit the deep semantics of source images, aiming to improve the quality of the fused image in a text-driven manner. However, challenges such as semantic noise and under-utilization of textual information still exist. Our approach effectively addresses these challenges by introducing entity-level semantics and multi-task learning, thereby improving semantic depth and the refinement of cross-modal interactions.

\subsection{Multi-task learning vision tasks}

Recent advances \cite{Mask,faster,facialexpression,videocaption,mtact,mtms} in multi-task learning  have demonstrated remarkable success in computer vision by enabling joint optimization of complementary tasks through shared feature representations. Mask R-CNN\cite{Mask} builds upon Faster R-CNN\cite{faster} by introducing a dedicated branch for semantic segmentation. This dual-branch structure allows for simultaneous bounding box and mask prediction, optimizing both tasks through shared features and improving overall efficiency and performance. By integrating  category label information and sample spatial distribution information, and introducing an adaptive reweighting module, DDMTL\cite{facialexpression} effectively improves the accuracy and robustness of facial expression recognition. MIMLDL\cite{videocaption} bridges the semantic gap between raw videos and generated captions by integrating a weakly supervised multi-label learning mechanism with an additional reconstructor.

Building upon these insights, our EGMT framework introduces multi-label classification as an auxiliary task,  bidirectionally transfers knowledge between the pixel-level fusion and entity-level recognition, where the classification task provides semantic constraints to guide feature integration while fused features enhance entity perception through enriched visual representations.

\subsection{Large vision-language models}

The advent of large vision-language models (LVLMs) has significantly improved the integration of multi-modal data, facilitating a more in-depth understanding of both visual and textual domains. These models excel in tasks such as image enhancement~\cite{image_enhancement,image_enhance_survey, single_image}, video processing~\cite{videocaption,video_cap,relative,triple_video} and visual question answering~\cite{BLIP2}. CLIP~\cite{CLIP}, a pioneering model, aligns visual and textual features in a joint embedding space by optimizing a contrastive loss function, enabling zero-shot performance on tasks such as image classification and image-text matching without fine-tuning. BLIP~\cite{blip} enhances CLIP by introducing a bootstrapping mechanism for better feature alignment, attaining leading results on visual question answering and image captioning through a tightly coupled vision encoder and language model framework. To reduce computational costs, BLIP-2\cite{BLIP2} further fixes N learnable query to represent image features and trains a lightweight Q-former to align visual-language features. LLaVA\cite{llava} further advances this by mapping image features to the language feature space using a linear projection layer, integrating image and text information effectively. 

In our EGMT, we use the widely used BLIP-2 to generate captions from source images. BLIP-2's capability to extract core descriptions from infrared and visible images, not only capturing key visual elements of the image, but also providing a basis for subsequent entity extraction.

\subsection{Entity extraction techniques}

Compared to the unprocessed text, entities can provide a structured representation of key information, effectively reducing the semantic noise inherent in original text\cite{TKDE,medical-text,text-based-person}. Entity extraction has been revolutionized by recent advances in pre-trained language models. FGET\cite{hou2021transfer} addresses label noise and feature representation limitations by leveraging sub-word embeddings, pre-trained language models, and topic models to generate efficient entity feature representations and refine predictions. GoLLIE\cite{gollie} proposes guideline-driven fine-tuning of large language models to achieve robust zero-shot generalization in open-domain entity extraction by explicitly aligning model predictions with annotation guidelines. Flan-T5\cite{FLAN-T5} as a novel instruction-tuned language model that enhances the generalization and adaptability of models, particularly in tasks such as entity extraction, by leveraging multi-task learning and zero/few-shot learning capabilities. 

Drawing from these insights, we employ a fine-tuned Flan-T5 model\cite{FACTUAL} as a dedicated entity extractor. This addresses the issue of semantic redundancy in the original text by filtering out irrelevant linguistic elements while preserving key semantic information.

\begin{figure*}[ht]
  \centering
  \includegraphics[width=\linewidth]{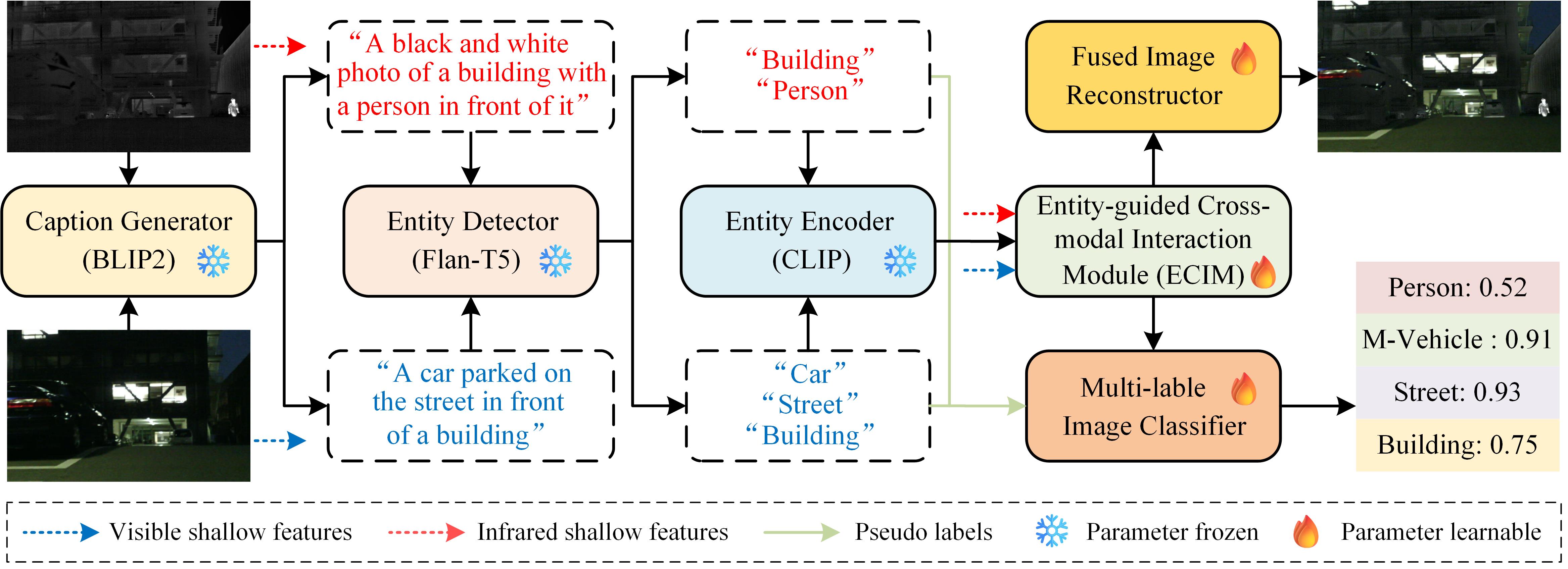}
  \caption{Workflow of our EGMT framework. Source images are processed by BLIP-2 to generate captions, from which Flan-T5 extracts entities that are encoded into features via CLIP. Concurrently, convolutional blocks extract shallow visual features from the source images. Both features are integrated through our entity-guided cross-modal interaction module (ECIM) to produce shared representations, which are finally decoded by task-specific heads into the fused image and classification results.}
  \label{fig:network}
\end{figure*}

\section{Method}~\label{sec:method}
This section details the overall architecture of EGMT (Sec.\ref{Overall Architecture}), encompassing the entity-guided cross-modal interaction module (Sec.~\ref{Entity-guided Cross-modal Interaction Module}), the task-specific heads(Sec.~\ref{Task-specific Heads}), and the design of the multi-task learning loss function(Sec.~\ref{Multi-task Learning Loss Function}).

\subsection{Overall architecture}~\label{Overall Architecture}
As depicted in Fig.~\ref{fig:network}, our model is composed of six main components: the caption generator, entity detector, entity encoder, entity-guided cross-modal interaction module, fused image reconstructor, and multi-label classifier. In detail, we first utilize BLIP-2\cite{BLIP2} to generate captions of the source images. Next, we leverage a pre-trained Flan-T5\cite{FACTUAL} to extract entities from the captions, and eliminate any duplicate entities through case-insensitive string matching. In practice, each image pair yields 2–8 entities, as dictated by its caption. The above operations constitute the entity extraction pipeline, which can to some extent remove the noise and hallucination issues from the raw captions. Then, these entities are input into the text encoder of CLIP ViT-L/14 to generate 768-dimensional entity features. Meanwhile, we utilize vanilla convolution blocks to extract the shallow visual features of the source images. Subsequently, inspired by the cross-modal interaction mechanism~\cite{cross_modal_1,cross_modal_2}, an entity-guided cross-modal interaction module (ECIM) is designed to facilitate inter-visual and visual-entity feature interaction, and enhance their semantic association. Finally, the output features after ECIM are input into task-specific heads to generate the fused image and multi-label classification results. Notably, for clearer expression, we simplify the subsequent processing flow of attention mechanisms, including the feed forward network and layer normalization.

\subsection{Entity-guided cross-modal interaction module}~\label{Entity-guided Cross-modal Interaction Module}
\begin{figure*}[ht]
  \centering
  \includegraphics[width=1\linewidth]{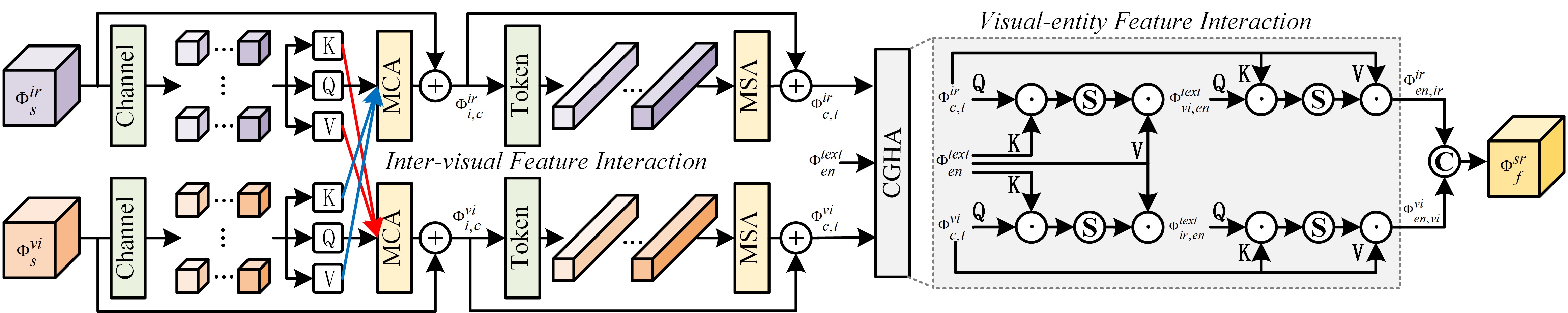}
  \caption{Architecture of the Entity-guided Cross-modal Interaction Module (ECIM). The module takes shallow visual features and entity features as inputs. The inter-visual interaction path (left) processes features through multi-head cross-attention (MCA) and multi-head self-attention (MSA) to capture intra- and inter-modal dependencies. The visual-entity interaction path (right) employs cross-modal guided hybrid attention (CGHA) to enable alternating guidance between visual and entity features.}
  \label{fig:module details}
\end{figure*}

Fig.~\ref{fig:module details} illustrates our utilization of the entity-guided cross-modal interaction module to capture the relationship of feature interaction across inter-visual and visual-entity modalities, which is beneficial to obtain a more comprehensive and semantically rich feature representation.

\subsubsection{Inter-visual feature interaction}

Given the shallow visual features \( {\Phi_{s}^{ir}} \) and \( {\Phi_{s}^{vi}} \), we first partition them into non-overlapping windows and compute query (\(Q\)), key (\(K\)), and value (\(V\)) vectors for each modality \( m \in \{ ir, vi \} \):

\begin{equation}
\left \{  {Q_{s,c}^{m},K_{s,c}^{m},V_{s,c}^{m}} \right \}=\left \{ \Phi_{s}^{m}W_{m}^{Q},\Phi_{s}^{m}W_{m}^{K},\Phi_{s}^{m}W_{m}^{V}    \right \}, 
\end{equation}

Next, across the channel axis, we can obtain the cross-channel interaction feature through multi-head cross-attention (\(MCA\)), which can be formulated as follows:

\begin{equation}
\Phi _{i,c}^{ir/vi}=MCA\left ( {Q}_{s,c}^{ir/vi}, K_{s,c}^{vi/ir} ,V_{i,c}^{vi/ir} \right ) + \Phi_{s}^{ir/vi}.
\end{equation}

Similarly, we embed \( \Phi _{i,c}^{ir} \) and \( \Phi _{s,c}^{vi} \) into a token-wise matrix \( M_t\in \mathbb{R}^{T\times L_t} \), where  \( T \) denotes the number of tokens and \( L_t \) denotes the length of the feature vector for each token. This allows us to obtain the corresponding \( Q_{i,t}^{m} \), \( K_{i,t}^{m} \), and \( V_{i,t}^{m} \).

Then, along the token dimension, we can obtain the token-wise global features \(\Phi _{c,t}^{m} \) (\(m \in \{ ir, vi \} \)) through multi-head self-attention (\(MSA\)), as follows:

\begin{equation}
\Phi _{c,t}^{m}=MSA\left ( Q_{i,t}^{m}, K_{i,t}^{m} ,V_{i,t}^{m} \right ) + \Phi_{i,c}^{m}.
\end{equation}

This inter-visual interaction mechanism, which integrates both channel and token dimensions, efficiently fuses multimodal visual features, enables in-depth semantic interaction across channels and spatial positions, and facilitates effective information exchange.

\subsubsection{Visual-entity feature interaction} To achieve the alternate guidance between visual and textual modalities, a cross-modal guided hybrid attention (CGHA) is proposed, we first apply a fully-connected layer to the entity features so that they share the same dimension as the visual features. Then, the visual features $\Phi_{c,t}^{m}$ are mapped to the query $Q_{c,t}^{m}$, and the aligned entity features to the key $K_{en}^{\text{text}}$ and value $V_{en}^{\text{text}}$. Guided by the image content itself, the scaled dot-product attention dynamically determines the weights of entity features, thereby effectively suppressing the interference of irrelevant entities (caused by LVLM hallucinations) on the visual features, which can be formulated as:

\begin{equation}
\left \{  {Q_{c,t}^{m},K_{en}^{text},V_{en}^{text}} \right \}=\left \{ \Phi_{c,t}^{m}W_{m}^{Q},\Phi_{en}^{text}W_{en}^{K},\Phi_{en}^{text}W_{en}^{V}  \right \}. 
\end{equation}

\begin{equation}
\Phi_{m,en}^{text} = \text{Softmax}\left(\frac{Q_{c,t}^{m} \cdot {K_{en}^{text}}^T}{\sqrt{d_k}}\right) \cdot V_{en}^{text}.
\end{equation}

Similarly, for the weighted entity features \(  \Phi_{m,en}^{text} \), we map them to the \( Q_{m,en}^{text}\), while the visual features \(  \Phi _{c,t}^{m} \)  are mapped to the  \( K_{c,t}^{m}\) and  \(V_{c,t}^{m}\) to leverage the entity information to enhance the semantic understanding of the image features, which can be expressed as follows:

\begin{equation}
\Phi_{en,m}^{m} = \text{Softmax}\left(\frac{Q_{m,en}^{text} \cdot {K_{c,t}^{m}}^T}{\sqrt{d_k}}\right) \cdot V_{c,t}^{m}.
\end{equation}

Through this visual-textual alternating guidance attention mechanism, the final features comprehensively reflect the intrinsic relationship between image content and relevant entity information, enabling a synergistic interaction between visual and textual modalities for enhanced semantic representation.

\subsection{Task-specific heads}~\label{Task-specific Heads}
Given the final shared representation features \( \Phi _{f}^{sr}\) obtained by ECIM, we enter them into task-specific heads to get the specific output for each task. 

For the image fusion task, we feed \( \Phi _{f}^{sr}\) into the reconstructor composed of three convolutional layers to obtain the fused results \({I_f}\), which can be expressed as:

\begin{equation}
{I_f} = Conv_1^{1}(Conv_1^{3}(Conv_1^{3}(\Phi _{f}^{sr}))),
\end{equation}
where the superscript and subscript of \( Conv \) indicate the kernel size and step size, respectively.

For the multi-label classification task, we embed \( {\Phi_{en}^{text}} \) into a visual-entity joint space. Subsequently, we apply a global max pooling operation and feed them into the fully connected layer to obtain the final classification results \( \hat{p} \), as follows:
\begin{equation}
 \hat{p}  = FC(MP({\Phi _{f}^{sr}\cdot{Mask({\Phi_{en}^{text}})}})),
\end{equation}
where \( FC\)  denotes the fully connected layer, \( MP\)  indicates the global max pooling operation, and \( Mask\) represents a uniform random masking operation with 60\% of the flattened entity features are zeroed out during training.

\subsection{Multi-task learning loss function}~\label{Multi-task Learning Loss Function}
In the proposed EGMT, our multi-task learning loss function is defined as follows:

\begin{equation}
{\mathcal{L}_{total}} = {\lambda_1}{\mathcal{L}_{fus}} + {\lambda_2}{\mathcal{L}_{cla}},
\end{equation}
where \(  {\lambda_1} \) and \(  {\lambda_2} \) control the balance between multi-task learning. \( {\mathcal{L}_{fus}} \) and \( {\mathcal{L}_{cla}} \) denote the loss functions for the image fusion task and the multi-label classification task, respectively.

The multi-task loss weights are adaptively determined through uncertainty learning\cite{dynaweight} :

\begin{equation}
\lambda_i = \frac{\exp(-w_i / \tau)}{\sum_{i=1}^2 \exp(-w_i / \tau)}, \quad i \in {1,2}
\end{equation}
where \( w_{i} \) denotes the original weight parameter of the \(i\)-th task, \( {\tau} \) represents the temperature coefficient, which is used to control the sharpness of the weights.

For the image fusion task, inspired by SwinFusion\cite{swinfusion}, the \( {L_{fus}} \)  is composed of intensity loss \( {L_{int}} \), edge loss \( {L_{edge}} \), and structural similarity loss \( {L_{ssim}} \), which can be expressed as:
\begin{equation}
{\mathcal{L}_{fus}} = {\alpha_1}{\mathcal{L}_{int}} + {\alpha_2}{\mathcal{L}_{edge}}+ {\alpha_3}{\mathcal{L}_{ssim}},
\end{equation}
where \(  {\alpha_1} \), \(  {\alpha_2} \) and \(  {\alpha _3} \) are the hyperparameters to control the trade-off.

The \( {\mathcal{L}_{int}} \), \( {\mathcal{L}_{edge}} \), and \( {\mathcal{L}_{ssim}} \) are expected to preserve the information of source images from different aspects, which can be expressed as:

\begin{equation}
{\mathcal{L}_{int}} = \frac{1}{{HW}}{\left\| {{I_f} - \max ({I_{ir}},{I_{vis}})} \right\|_1},
\end{equation}

\begin{equation}
{\mathcal{L}_{edge}} = \frac{1}{{HW}}{\left\| {|\nabla {I_f}| - \max (|\nabla {I_{ir}}|,|\nabla {I_{vis}}|)} \right\|_1},
\end{equation}

\begin{equation}
{\mathcal{L}_{ssim}} = (1-SSIM({I_f},{I_{ir}})) + (1-SSIM({I_f},{I_{vis}})),
\end{equation}
where \( \max ( \cdot ) \) denotes the element-wise maximum selection, \( {\left\|  \cdot  \right\|_1} \) denotes the L1-norm, and \( \nabla \) denotes the Laplacian operator.

For the multi-label classification task, limited by the scenarios of the MSRS dataset, we adopt the focal loss\cite{focal} to solve the issue of class imbalance, which is expressed as:

\begin{equation}
\mathcal{L}_{cla} = \frac{1}{N} \sum_{i=1}^{N} \alpha_i \cdot \left(1 - \hat{p}_i\right)^\gamma \cdot \text{BCE}(p_i, y_i),
\end{equation}
where \(N\) indicates the number of samples for the class, \(\alpha_i\) is the class-balancing weight inversely proportional to the positive class sample ratio, \(\gamma\) is the focusing parameter, and \(\text{BCE}(\cdot)\) denotes the binary cross-entropy loss. The prediction probability \(p_i \in [0, 1]\) is the model's predicted probability for the \(i\)-th category, and the true label \(y_i \in \{0, 1\}\) is the ground truth label for the \(i\)-th category. 

\begin{figure*}[!ht]
  \centering
  \includegraphics[width=\linewidth]{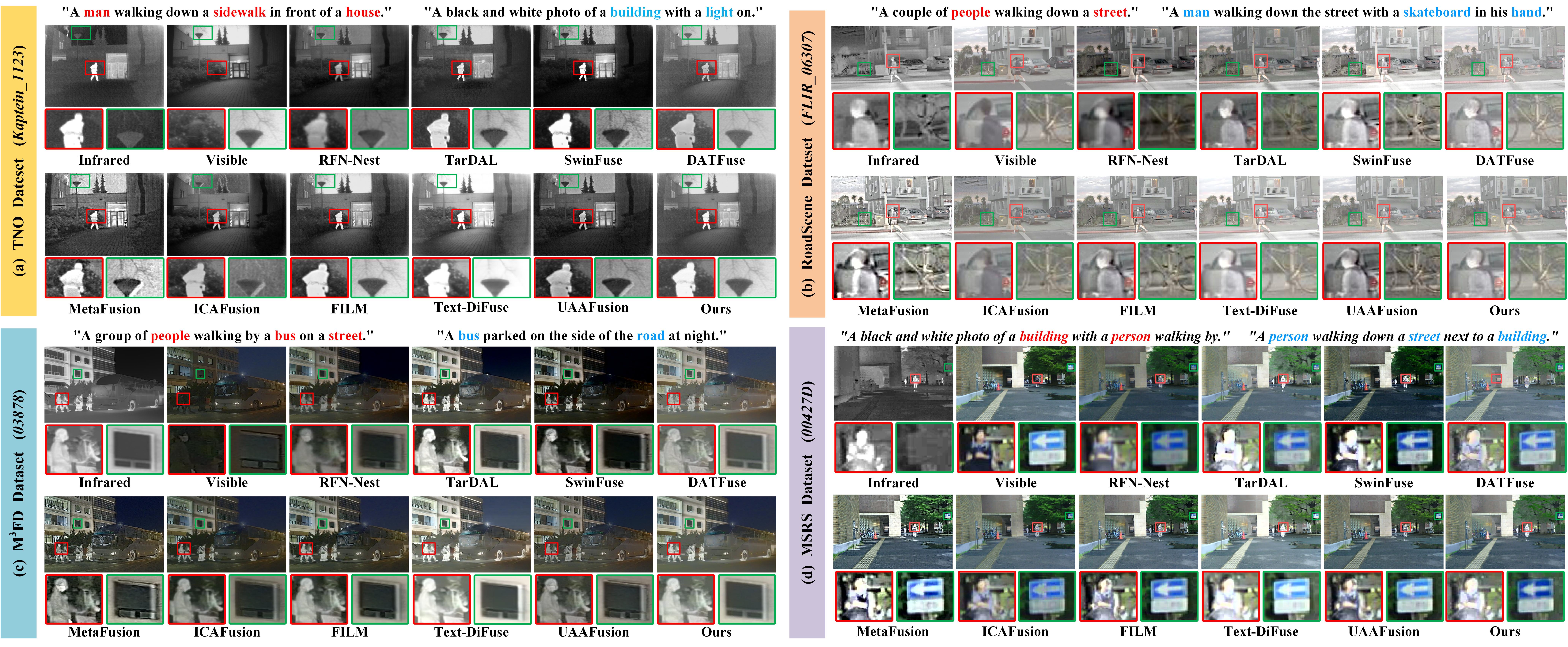}
\caption{Qualitative assessment of our EGMT and the other SOTA methods across TNO, RoadScene, M\(^{3}\)FD, and MSRS datasets. Specifically, four representative examples, namely \textit{Kaptein\_1123}, \textit{FILR\_06307}, \textit{03878}, and \textit{00427D} are selected for visual comparison. Entities from the infrared image captions are rendered in red, and those from the visible image captions are in blue. 
}
  \label{fig:visual}
\end{figure*}

\section{Experiments}~\label{sec:experiment}
This section starts with an overview of our experimental setup. Then, we analyze the experimental comparisons, which include image fusion, multi-label classification, downstream application, and efficiency assessment, to verify the effectiveness of our proposed EGMT. Finally, we discuss the results of the ablation study and the limitations of our EGMT.

\subsection{Experimental setup}

\subsubsection{Dataset} In the training phase, we crop the images to 448 × 448 sizes with a fixed sliding step from the MSRS\cite{PIAFusion} training set, which includes 1086 image pairs, and obtain a training set consisting of 10563 patch-level image pairs. In the testing phase, we utilize the TNO\cite{TNO} dataset (25 pairs of images), the MSRS test set (361 pairs of images), the M\(^{3}\)FD\cite{TAR} dataset (300 pairs of independent scenes for fusion), the RoadScene\cite{fusiondn} dataset (221 pairs of images), and their corresponding entity features to conduct a comprehensive evaluation in the image fusion task. Meanwhile, we evaluate the performance of the multi-label classification task on the MSRS test set. In addition, the downstream task is performed on the M\(^{3}\)FD dataset.

\subsubsection{Comparison Methods \& Evaluation Metrics} For the image fusion task, we select nine cutting-edge image fusion methods for comparison, which include RFN-Nest~\cite{rfn}, TarDAL~\cite{TAR}, SwinFuse~\cite{swin}, DATFuse~\cite{DATFuse}, MetaFusion~\cite{meta}, ICAFusion~\cite{ICAF}, FILM~\cite{FILM}, Text-DiFuse~\cite{Text-diffuse}, and UAAFusion~\cite{UAA}. For a thorough assessment, we utilize eight metrics to evaluate the fusion performance. These metrics encompass mutual information (MI), phase congruency (PC), structural similarity index measure (SSIM), peak signal-to-noise ratio (PSNR), normalized absolute bilateral filtering error (N$_{abf}$), gradient-based similarity measurement (Q$_{abf}$), nonlinear correlation information entropy (NCIE), and feature mutual information based on wavelet (FMI$_{w}$). For the multi-label classification task, we adopt six metrics to verify the classification performance of our proposed EGMT, including hamming loss (HL), ranking loss (RL), mean average precision (mAP), area under the receiver operating characteristic curve (AUC), jaccard index (JI), and F1 score (F1).

\subsubsection{Implementation Details} We use the Adam optimizer to train the model, with the learning rate set to \( 1 \times 10^{-4} \). The batch size and number of epochs are set to 4 and 20, respectively. The non-overlapping patch size is 16 × 16. The number of channels for shallow visual features is 32, and the input-output channel configurations for the convolutional layers in the fused image reconstructor are 64:32, 32:16, and 16:1, respectively. In our loss function, the hyperparameters  \( {\alpha _1} \), \( {\alpha _2} \), and \( {\alpha _3} \) are set to 1, 15, and 5, respectively. The focusing parameter \( \gamma \) is set to 2 and the temperature coefficient \( {\tau} \) is fixed at 1. The experimental setup is powered by an NVIDIA GeForce RTX 4090 GPU.

\subsection{Comparative analysis}

\subsubsection{Fusion results on TNO dataset}

\begin{table*}[!htbp]
	\renewcommand\arraystretch{1.5}
	\caption{Quantitative assessment of our EGMT and the other SOTA methods on the TNO dataset. (MT: multi-task learning; TI: text interaction; \textcolor{red}{\textbf{Red}}: optimal; \textcolor{blue}{\textit{Blue}}: sub-optimal)}
	\centering
	\begin{tabular}{ l c c c c c c c c c c c }
		\hline
        Methods & Publishers& MT & TI & PC ${\uparrow}$ & SSIM ${\uparrow}$ & MI ${\uparrow}$ & Q$_{abf}$ ${\uparrow}$  & PSNR ${\uparrow}$ & FMI$_{w}$ ${\uparrow}$ & N$_{abf}$ ${\downarrow}$  & NCIE ${\uparrow}$   \\
		\hline
		RFN-Nest\cite{rfn}& Inf. Fus.$'$21& \XSolidBrush & \XSolidBrush  & 0.23990 & 0.69542 & 1.92851 & 0.35442 & 15.39564 & 0.30799 & 0.01785 & 0.80428	\\
		TarDAL\cite{TAR}& CVPR$'$22 & \Checkmark & \XSolidBrush & 0.24497 & \textcolor{red}{\bfseries0.72640}  & 2.63654& 0.40747 & \textcolor{red}{\bfseries16.58085} & 0.35941 & 0.03433 & 0.80628       \\
		SwinFuse\cite{swin}& TIM$'$22 & \XSolidBrush & \XSolidBrush &0.28554 & 0.65526 & 2.32333 & 0.43253 & 14.84917 & 0.42623 & 0.07300 & 0.80534   \\
		DATFuse\cite{DATFuse} & TCSVT$'$23 & \XSolidBrush & \XSolidBrush & 0.36684 & 0.70445 & 3.23802 & 0.50429 & 15.70604 & 0.41760 & \textcolor{blue}{\itshape0.01392} & 0.80810 \\
		MetaFusion\cite{meta}& CVPR$'$23 & \Checkmark & \XSolidBrush &0.25034 & 0.54538 & 1.54185 & 0.31308 & 15.19279 & 0.36710 & 0.29305 & 0.80361		\\
        ICAFusion\cite{ICAF}& TMM$'$23 & \XSolidBrush & \XSolidBrush & 0.35081 & 0.63429 & \textcolor{blue}{\itshape3.74249} & 0.46750 & 15.55394 & \textcolor{blue}{\itshape0.44885} & 0.08876 & \textcolor{blue}{\itshape0.81145} 	 \\
        FILM\cite{FILM}& ICML$'$24 & \XSolidBrush  & \Checkmark & 0.33621 & 0.67355 & 2.91080 & 0.51317 & 15.81941 & 0.37431 & 0.07222 & 0.80755		\\
        Text-DiFuse\cite{Text-diffuse}& NeurIPS$'$24 & \XSolidBrush  & \Checkmark & 0.27034 &0.68017 &2.84842 &0.43652 &13.24773 &0.25802 &0.03873 &0.80685 \\
        UAAFusion\cite{UAA}& TCSVT$'$25 & \XSolidBrush  & \XSolidBrush & \textcolor{blue}{\itshape0.36908} &0.70264 &1.92674 &\textcolor{blue}{\itshape0.55679} &16.37341 &0.40704 &0.06320 &0.80436 \\

		Ours& -- & \Checkmark & \Checkmark & \textcolor{red}{\bfseries0.40241} & \textcolor{blue}{\itshape0.72085} & \textcolor{red}{\bfseries4.86354} & \textcolor{red}{\bfseries0.56391} & \textcolor{blue}{\itshape16.46613} & \textcolor{red}{\bfseries0.45979} & \textcolor{red}{\bfseries0.00595} & \textcolor{red}{\bfseries0.81674} \\
		\hline 		
        \label{tab:TNO}
	\end{tabular}
\end{table*}

The qualitative assessment is shown in Fig.~\ref{fig:visual}~(a), analyzing the results, we observe that RFN-Nest, SwinFuse, DATFuse, and ICAFusion have problems of brightness and detail loss. The approaches focused on semantic-level, such as TarDAL, MetaFusion and UAAFusion, while preserving the target brightness, introduce some noise in the fusion results due to the involvement of high-level tasks. FILM slightly lower target brightness than our method, but the overall image is visually similar. Text-DiFuse, a text-guided diffusion-based method, achieves the most favorable target brightness. However, the high-frequency details are lost in the denoising process. 

Table~\ref{tab:TNO} presents the quantitative comparison results of EGMT with the other state-of-the-art (SOTA) methods. Our EGMT ranks first in PC, MI, Q$_{abf}$, FMI$_{w}$, N$_{abf}$, and NCIE, and second in SSIM and PSNR. The best MI and NCIE reveal that the fused images retain more meaningful information. The best Q$_{abf}$ and FMI$_{w}$ indicate the superior preservation of edge information. The optimal PC and N$_{abf}$ demonstrate that our EGMT successfully preserves crucial structural details and achieves superior visual outcomes while minimizing the presence of artifacts. The suboptimal SSIM and PSNR values reveal that our EGMT is effective in preserving the structural integrity of the source image while effectively suppressing noise.

\begin{table*}[!t]
	\renewcommand\arraystretch{1.5}
	\caption{Quantitative assessment of our EGMT and the other SOTA methods on the RoadScene dataset. (MT: multi-task learning; TI: text interaction; \textcolor{red}{\textbf{Red}}: optimal; \textcolor{blue}{\textit{Blue}}: sub-optimal)}
	\centering
	\begin{tabular}{ l c c c c c c c c c c c }
		\hline
        Methods & Publishers& MT & TI & PC ${\uparrow}$ & SSIM ${\uparrow}$ & MI ${\uparrow}$ & Q$_{abf}$ ${\uparrow}$  & PSNR ${\uparrow}$ & FMI$_{w}$ ${\uparrow}$ & N$_{abf}$ ${\downarrow}$  & NCIE ${\uparrow}$   \\
		\hline
		RFN-Nest\cite{rfn}& Inf. Fus.$'$21& \XSolidBrush & \XSolidBrush  & 0.26552 & 0.65552 & 2.77585 & 0.31308 & 13.98696 & 0.27533 & 0.00758 & 0.80649	\\
		TarDAL\cite{TAR}& CVPR$'$22 & \Checkmark & \XSolidBrush & 0.31821 & 0.68793  & 3.47019 & 0.43402 & 15.05293 & 0.33408 & 0.02287 & 0.80873      \\
		SwinFuse\cite{swin}& TIM$'$22 & \XSolidBrush & \XSolidBrush & 0.38584 & 0.67781 & 3.17831 & 0.50671 & 14.30816 & \textcolor{blue}{\itshape0.43612} & 0.05464 & 0.80791   \\
		DATFuse\cite{DATFuse} & TCSVT$'$23 & \XSolidBrush & \XSolidBrush & 0.38921 & 0.69458 & 3.78383 & 0.48366 & \textcolor{blue}{\itshape15.78547} & 0.38223 & \textcolor{blue}{\itshape0.00591} & 0.80971 \\
		MetaFusion\cite{meta}& CVPR$'$23 & \Checkmark & \XSolidBrush & 0.29299 & 0.57802 & 2.22826 & 0.38227 & 13.84251 & 0.37391 & 0.20879 & 0.80518		\\
        ICAFusion\cite{ICAF}& TMM$'$23 & \XSolidBrush & \XSolidBrush & 0.37959 & 0.62783 & \textcolor{blue}{\itshape3.84998} & 0.40410 & 13.33192 & 0.40501 & 0.05176 & \textcolor{blue}{\itshape0.81057} 	 \\
        FILM\cite{FILM}& ICML$'$24 & \XSolidBrush  & \Checkmark & 0.41014 & 0.66544 & 3.09260 & \textcolor{red}{\bfseries0.58625} & 15.07319 & 0.38576 & 0.05675 & 0.80753		\\
        Text-DiFuse\cite{Text-diffuse}& NeurIPS$'$24 & \XSolidBrush  & \Checkmark &0.39407 &\textcolor{blue}{\itshape0.70123} &3.54959 &0.44810 &14.19185 &0.32391 &0.02813 &0.80883 \\
        UAAFusion\cite{UAA}& TCSVT$'$25 & \XSolidBrush  & \XSolidBrush & \textcolor{blue}{\itshape0.41941} &0.68247 &2.51909 &\textcolor{blue}{\itshape0.57118} &15.59690 &0.37183 &0.05850 &0.80588 \\

		Ours& -- & \Checkmark & \Checkmark & \textcolor{red}{\bfseries0.43860} & \textcolor{red}{\bfseries0.70555} & \textcolor{red}{\bfseries5.42253} & 0.52848 & \textcolor{red}{\bfseries16.13816} & \textcolor{red}{\bfseries0.44724} & \textcolor{red}{\bfseries0.00440} & \textcolor{red}{\bfseries0.81806} \\
		\hline 		
        \label{tab:road}
	\end{tabular}
\end{table*}

\subsubsection{Fusion results on RoadScene dataset}

We further conduct qualitative and quantitative analyses on the RoadScene dataset to validate the fusion performance of EGMT. Fig.~\ref{fig:visual}(b) presents the qualitative results. In terms of  texture structure, EGMT preserves finer details than competing methods. In the pedestrian region, EGMT maintains a balanced brightness control, slightly behind SwinFuse, MetaFusion and Text-DiFuse. SwinFuse adopts an L1-norm fusion strategy that preserves overall brightness but sacrifices fine details due to the sparsity inherent in L1-regularized solutions. MetaFusion projects semantic features from the object-detection task and fusion features into a shared latent space, which inevitably erodes structural details.

Table~\ref{tab:road} presents the quantitative assessment results. From these results, our EGMT ranks first in PC, SSIM, MI, PSNR, FMI$_{w}$, N$_{abf}$, and NCIE. Moreover, for the metric Q$_{abf}$, EGMT places third, slightly trailing behind FILM and UAAFusion. Through a comprehensive metric evaluation, our EGMT shows its overall superiority, which is consistent with the results of qualitative evaluation.

\begin{table*}[!ht]
	\renewcommand\arraystretch{1.5}
        \caption{Quantitative assessment of our EGMT and the other SOTA methods on the M\(^{3}\)FD dataset. (MT: multi-task learning; TI: text interaction; \textcolor{red}{\textbf{Red}}: optimal; \textcolor{blue}{\textit{Blue}}: sub-optimal)}
	\centering
	\begin{tabular}{ l c c c c c c c c c c c}
		\hline
        Methods & Publishers& MT & TI & PC ${\uparrow}$ & SSIM ${\uparrow}$ & MI ${\uparrow}$ & Q$_{abf}$ ${\uparrow}$  & PSNR ${\uparrow}$ & FMI$_{w}$ ${\uparrow}$ & N$_{abf}$ ${\downarrow}$  & NCIE ${\uparrow}$   \\
		\hline
		RFN-Nest\cite{rfn}& Inf. Fus.$'$21 & \XSolidBrush & \XSolidBrush & 0.31198 & \textcolor{red}{\bfseries0.72843} & 2.87861 & 0.40595 & 15.55149 & 0.28167 & \textcolor{blue}{\itshape0.00302} & 0.80725	\\
		TarDAL\cite{TAR}& CVPR$'$22& \Checkmark & \XSolidBrush & 0.28897 & 0.70765 & 3.17326 & 0.41932 & 15.30518 & 0.33474 & 0.03045 & 0.80827      \\
		SwinFuse\cite{swin}& TIM$'$22 & \XSolidBrush & \XSolidBrush & 0.39365 & 0.68038 & 3.17525 & 0.53842 & 14.66071& \textcolor{blue}{\itshape0.41343} & 0.04098 & 0.80824   \\
        DATFuse\cite{DATFuse} & TCSVT$'$23 & \XSolidBrush & \XSolidBrush & 0.36466 & 0.70408 & 4.12986 & 0.49356 & 15.68320 & 0.39336 & 0.00618 & 0.81200 \\
        MetaFusion\cite{meta}& CVPR$'$23& \Checkmark & \XSolidBrush & 0.37244 & 0.60962 & 2.35203 & 0.39226 & 14.50725 & 0.32564 & 0.24918 & 0.80570		\\
		ICAFusion\cite{ICAF}& TMM$'$23& \XSolidBrush & \XSolidBrush & 0.41829& 0.69443 & 3.99578 & 0.56469 & 15.33982 & 0.40794 & 0.03594 & 0.81111 	 \\
        FILM\cite{FILM}& ICML$'$24& \XSolidBrush  & \Checkmark & \textcolor{blue}{\itshape0.43857} & 0.70922 & \textcolor{blue}{\itshape4.39459} & \textcolor{blue}{\itshape0.63834} & \textcolor{red}{\bfseries16.79113} & 0.39850 & 0.03563 & \textcolor{blue}{\itshape0.81479}		\\
        Text-DiFuse\cite{Text-diffuse}& NeurIPS$'$24 & \XSolidBrush  & \XSolidBrush & 0.32311 &0.65624 &3.30867 &0.49202 &12.21278 &0.24330 &0.04213 &0.80838 \\
        UAAFusion\cite{UAA}& TCSVT$'$25 & \XSolidBrush  & \XSolidBrush & \textcolor{red}{\bfseries0.52229} &\textcolor{blue}{\itshape0.72728} &2.72229 &\textcolor{red}{\bfseries0.68036} &16.08453 &0.39501 &0.02563 &0.80672 \\
		Ours& --& \Checkmark & \Checkmark & 0.42021  & 0.71954 & \textcolor{red}{\bfseries5.56808} & 0.55580 & \textcolor{blue}{\itshape16.47044} & \textcolor{red}{\bfseries0.46910} & \textcolor{red}{\bfseries0.00288} & \textcolor{red}{\bfseries0.82122} \\
		\hline 		
        \label{tab:m3fd}
	\end{tabular}
\end{table*}

\begin{table*}[ht]
	\renewcommand\arraystretch{1.5}
        \caption{Quantitative assessment of our EGMT and the other SOTA methods on the MSRS dataset. (MT: multi-task learning; TI: text interaction; \textcolor{red}{\textbf{Red}}: optimal; \textcolor{blue}{\textit{Blue}}: sub-optimal)}
	\centering
	\begin{tabular}{ l c c c c c c c c c c c}
		\hline
        Methods & Publishers& MT & TI & PC ${\uparrow}$ & SSIM ${\uparrow}$ & MI ${\uparrow}$ & Q$_{abf}$ ${\uparrow}$  & PSNR ${\uparrow}$ & FMI$_{w}$ ${\uparrow}$ & N$_{abf}$ ${\downarrow}$  & NCIE ${\uparrow}$   \\
		\hline
		RFN-Nest\cite{rfn}& Inf. Fus.$'$21& \XSolidBrush & \XSolidBrush  & 0.32778 & 0.67110 & 3.32270 & 0.24567 & \textcolor{red}{\bfseries23.98411} & 0.27532 & \textcolor{blue}{\itshape0.00530} & 0.81051	\\
		TarDAL\cite{TAR}& CVPR$'$22& \Checkmark & \XSolidBrush & 0.31079 & 0.52018 & 2.64606 & 0.42577 & 14.41296 & 0.34597 & 0.01936 & 0.80695       \\
		SwinFuse\cite{swin}& TIM$'$22  & \XSolidBrush & \XSolidBrush & 0.34452 & 0.61476 & 2.93857 & 0.45707 & 18.74044 & 0.29473 & 0.00571 & 0.80829    \\
		DATFuse\cite{DATFuse} & TCSVT$'$23 & \XSolidBrush & \XSolidBrush & 0.49181 & 0.62654 & 3.89000 & 0.64025 & 17.86303 & 0.41172 & 0.01340 & 0.81285 \\
        MetaFusion\cite{meta}& CVPR$'$23& \Checkmark & \XSolidBrush & 0.33600 & 0.63224 & 1.52687 & 0.45614 & 17.64087 & 0.31700 & 0.13014 & 0.80334		\\
		ICAFusion\cite{ICAF}& TMM$'$23& \XSolidBrush & \XSolidBrush & 0.34682 & 0.63485 & 3.66218 & 0.41203 & 18.22453 & 0.34179 & 0.01521 & 0.81160 	 \\
        FILM\cite{FILM}& ICML$'$24 & \XSolidBrush  & \Checkmark & \textcolor{blue}{\itshape0.53893} & 0.70697 & \textcolor{blue}{\itshape4.95116} & \textcolor{red}{\bfseries0.72639} & \textcolor{blue}{\itshape20.82279} & \textcolor{blue}{\itshape0.43768} & 0.02235 & \textcolor{blue}{\itshape0.82313}		\\
        Text-DiFuse\cite{Text-diffuse}& NeurIPS$'$24 & \XSolidBrush  & \Checkmark & 0.28248 &0.51003 &2.56881 &0.43941 &12.51039 &0.23590 &0.07458 &0.80622 \\
        UAAFusion\cite{UAA}& TCSVT$'$25 & \XSolidBrush  & \XSolidBrush & 0.45597 &\textcolor{red}{\bfseries0.72042} &2.30164 &0.62956 &18.93841 &0.35553 &0.01698 &0.80568 \\
		Ours& --& \Checkmark & \Checkmark & \textcolor{red}{\bfseries0.54329}  & \textcolor{blue}{\itshape0.71117} & \textcolor{red}{\bfseries5.73039} & \textcolor{blue}{\itshape0.69927} & 20.75212 & \textcolor{red}{\bfseries0.46649} & \textcolor{red}{\bfseries0.00125} & \textcolor{red}{\bfseries0.82777} \\
		\hline 		
        \label{tab:msrs}
	\end{tabular}
\end{table*}

\subsubsection{Fusion results on M\(^{3}\)FD dataset}
We extend our evaluation by performing experiments on the M\(^{3}\)FD dataset, thoroughly comparing EGMT with other competitors to assess its generalizability. Fig.~\ref{fig:visual} (c) shows the qualitative assessment results. In terms of background details, methods such as RFN-Nest, Tardal, SwinFuse, MetaFusion, ICAFusion, FILM, Text-DiFuse, and UAAFusion fail to effectively transfer the texture information of visible images into the fused results. Meanwhile, in the pedestrian region, our EGMT remains competitive. Tardal, MetaFusion, and Text-DiFuse achieve higher target brightness by constraining the image fusion task through object detection, but at the cost of sacrificing background details. Overall, EGMT achieves more balanced fusion results.

For the quantitative comparison results, as shown in Table~\ref{tab:m3fd}, our EGMT obtains the optimal values for MI, FMI$_{w}$, N$_{abf}$, and NCIE. Moreover, for PC and SSIM, our EGMT places third. Through both qualitative and quantitative analysis, we show that entity-guided cross-modal interaction can offer valuable complementary information, enhancing the performance of our fusion model.

\begin{figure*}[ht]
  \centering
  \includegraphics[width=\linewidth]{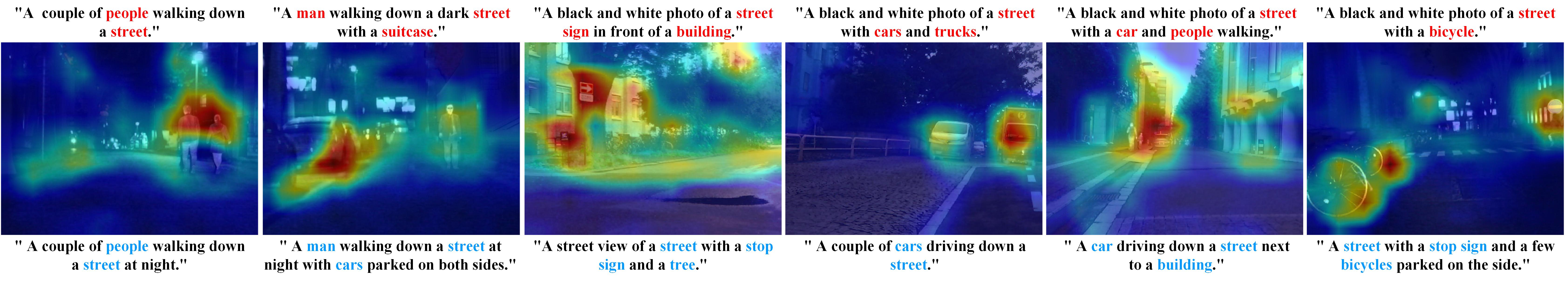}
\caption{ Entity activation heat maps of our EGMT on the MSRS dataset.}
  \label{fig:cam}
\end{figure*}

\begin{table}[ht]
\renewcommand\arraystretch{1.5}
\centering
\caption{Multi-label classification performance of EGMT framework with different input modalities on the MSRS dataset. }
\label{tab:mlc}
\begin{tabular}{ l c c c c c c }
\toprule
Input Modality & HL ${\downarrow}$ & RL ${\downarrow}$  & mAP${\uparrow}$  & AUC ${\uparrow}$  & J1 ${\uparrow}$  & F1 ${\uparrow}$ \\
\midrule
Only\_IR & 0.229 & 0.187 & 0.513 & 0.674 & 0.432 & 0.603 \\
Only\_VIS & 0.207 & 0.165 & 0.612 & 0.738 & 0.480 & 0.649\\
Fused (IR+VIS) & 0.184 & 0.136 & 0.779 & 0.857 & 0.554 &0.713 \\
\bottomrule
\end{tabular}
\end{table}

\subsubsection{Fusion and classification results on MSRS 
dataset}
We evaluate the fusion performance on the MSRS test set to verify the robustness of our EGMT. Fig.~\ref{fig:visual} (d) presents the qualitative results. Compared with other competitors, our fused image achieves a significantly better balance between thermal targets and texture details, which is consistent with our previous analysis. This is because guiding cross-modal interaction with entity information enables the fused image to focus more effectively on the distinctive features of both infrared and visible images. Moreover, Table~\ref{tab:msrs} shows the quantitative comparison results, our EGMT ranks first in PC, MI, FMI$_{w}$, N$_{abf}$, and NCIE, and second in SSIM and Q$_{abf}$. These rankings further highlight the effectiveness of our EGMT in generating high-quality fused images.

Our EGMT combines image fusion with multi-label classification, establishing entity-guided semantic constraints through nine categorical pseudo-labels. Table~\ref{tab:mlc} shows the multi-label classification performance of our EGMT. The HL of 0.184 indicates that 81.6\% of the labels are correctly predicted, which suggests that our EGMT is generally effective in assigning labels correctly in multi-label classification. The mAP of 0.779 demonstrates strong classification performance across different classes. The AUC score of 0.858 indicates effective discrimination between positive and negative classes. The JI of 0.554 suggests that the EGMT achieves reasonably accurate predictions. Notably, the multimodal inputs significantly outperformed the unimodal baseline on all evaluation metrics, empirically validating the semantic superiority of fused images. Furthermore, as shown in Fig.~\ref{fig:cam}, our EGMT framework effectively highlights critical entity regions in fused images, which further proves that multi-label classification improves visual features by guiding the better alignment of modality-aware representations.

\subsection{Downstream application}

\begin{figure*}[ht]
  \centering
  \includegraphics[width=\linewidth]{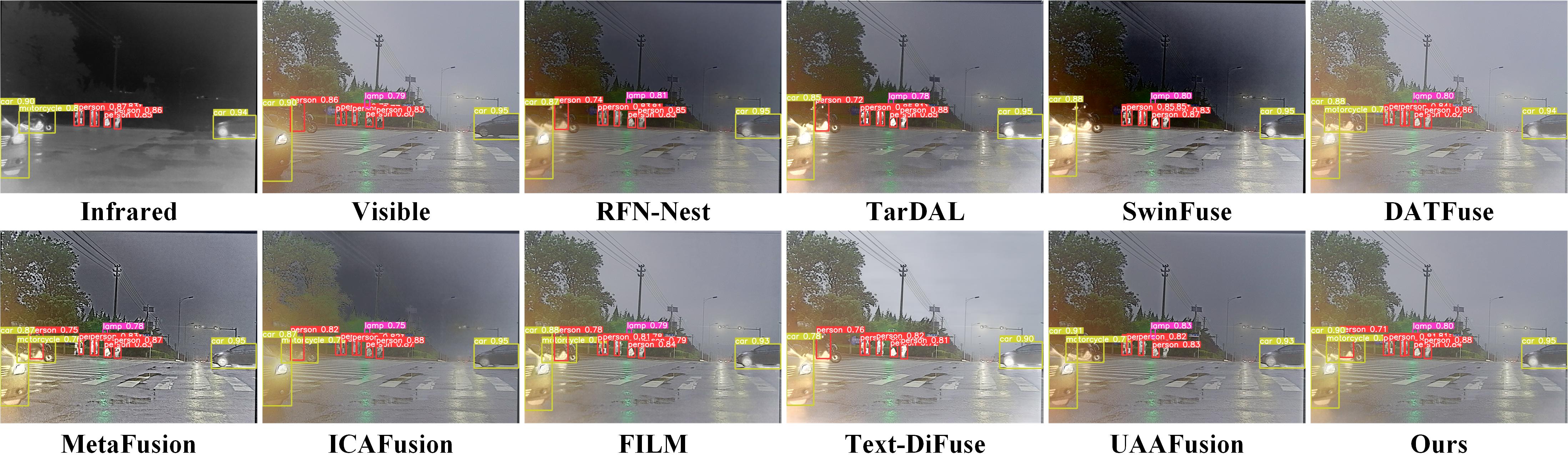}
\caption{Qualitative assessment of the detection performance of our EGMT and the other SOTA methods on the M$ ^3 $FD dataset.}
  \label{fig:object}
\end{figure*}

\begin{table*}[ht]
	\renewcommand\arraystretch{1.5}
	\renewcommand\tabcolsep{6.5pt}
	\scriptsize
	\caption{Quantitative assessment of the detection performance of our EGMT and the other SOTA methods on the M$ ^3 $FD dataset. (\textcolor{red}{\textbf{Red}}: optimal; \textcolor{blue}{\textit{Blue}}: sub-optimal)}
	\label{object}
	\centering

	\begin{tabular}{l |c c c c c c c | c c c c c c c}
		\hline
		\multirow{2}{*}{Methods}  &\multicolumn{7}{|c|}{mAP@0.5}	&\multicolumn{7}{c}{mAP@[0.5:0.95]}  \\
		\cline{2-15}				
		& Person & Car & Bus & Lamp & Motorcycle & Truck & All & Person & Car & Bus & Lamp & Motorcycle & Truck  & All \\
		\hline
		Infrared & \textcolor{red}{\bfseries0.781} & 0.868 &  0.924 & 0.659 & 0.766 & 0.856 & 0.809
		& \textcolor{red}{\bfseries0.542} & 0.671 & 0.777 & 0.358 &  0.513 & 0.673 & 0.589 \\
		
		Visible	& 0.719 &  0.890 & 0.929 & 0.782 & \textcolor{red}{\bfseries0.785} & 0.866 & 0.828	 
        & 0.477 & 0.699 & 0.793 & \textcolor{red}{\bfseries0.460} & 0.531 & 0.696 & 0.609 \\
		
		RFN-Nest \cite{rfn}  
		& 0.767 & \textcolor{red}{\bfseries0.892} & 0.935 & \textcolor{blue}{\itshape0.785} & 0.779 & \textcolor{blue}{\itshape0.875} & \textcolor{blue}{\itshape0.839}
		& 0.531 & \textcolor{blue}{\itshape0.705} & \textcolor{blue}{\itshape0.797} & 0.455 & 0.538 & 0.697 & 0.621 \\
		
		TarDAL \cite{TAR}  
		& 0.777 & 0.885 & 0.926 & 0.771 & 0.783 & 0.865 & 0.835
		& \textcolor{blue}{\itshape0.541} & 0.698 & 0.792 & 0.454 & 0.539 & 0.688 & 0.619 \\
		
		SwinFuse \cite{swin}
		& 0.778 & \textcolor{blue}{\itshape0.891} & 0.931 & 0.782 & 0.774 & 0.871 & 0.838
		& 0.539 & 0.704 & 0.795 & \textcolor{blue}{\itshape0.459} & \textcolor{blue}{\itshape0.548} & \textcolor{blue}{\itshape0.698} & \textcolor{red}{\bfseries0.624} \\
		
		DATFuse \cite{DATFuse} 
		& 0.772 & 0.881 & 0.929 & 0.758 & 0.762 & 0.870 & 0.829
		& 0.532 & 0.694 & 0.788 & 0.442 & 0.521 & 0.691 & 0.611 \\
		
		MetaFusion \cite{meta}  
		& 0.773 & 0.889 & 0.934 & 0.778 & \textcolor{red}{\bfseries0.785} & \textcolor{blue}{\itshape0.875} & \textcolor{blue}{\itshape0.839}
		& 0.532 & 0.702 & 0.795 & 0.456 & 0.538 & 0.695 & 0.620  \\
		
		ICAFusion \cite{ICAF}
		& 0.759 & 0.887 & 0.935 & 0.773 & 0.778 & 0.870 & 0.834
		& 0.519 & 0.699 & 0.793 & 0.446 & \textcolor{red}{\bfseries0.554} & 0.692 & 0.617 \\
		
		FILM \cite{FILM}
		& 0.777 & 0.889 & \textcolor{blue}{\itshape0.937} & \textcolor{red}{\bfseries0.786} & 0.753 & 0.862 & 0.834	
		& \textcolor{red}{\bfseries0.542} & 0.703 & \textcolor{blue}{\itshape0.797} & 0.454 & 0.535 & 0.688 & 0.620 \\
        Text-DiFuse\cite{Text-diffuse} &\textcolor{blue}{\itshape0.780} &0.889 &0.928 &0.774 &\textcolor{blue}{\itshape0.784} &0.870 &0.838 &  0.539 &0.703 &0.796 &0.446 &0.539 &0.686 &0.618 \\
        UAAFusion\cite{UAA} & 0.774 &\textcolor{red}{\bfseries0.892} &\textcolor{red}{\bfseries0.938} &\textcolor{red}{\bfseries0.786} &0.779 &0.877 &0.841 &0.536 &\textcolor{red}{\bfseries0.706} &\textcolor{blue}{\itshape0.797} &0.458 &0.539 &0.697 &\textcolor{blue}{\itshape0.622} \\	
		Ours  
		& \textcolor{blue}{\itshape0.780} & \textcolor{blue}{\itshape0.891} & \textcolor{red}{\bfseries0.938} & \textcolor{blue}{\itshape0.785} & \textcolor{blue}{\itshape0.784} & \textcolor{red}{\bfseries0.882} & \textcolor{red}{\bfseries0.843}
		& \textcolor{blue}{\itshape0.541} & 0.702 & \textcolor{red}{\bfseries0.805} & \textcolor{red}{\bfseries0.460} & 0.540 & \textcolor{red}{\bfseries0.699} & \textcolor{red}{\bfseries0.624} \\
		
		\hline
	\end{tabular}
\end{table*}

Object detection is one of the most widely used downstream tasks. By fully utilizing the texture details and thermal radiation information from the source images, the fused image overcomes the limitations of the single RGB mode, which significantly improves the accuracy and robustness of object detection. In this paper, we select the YOLOv5 model\cite{yolo} as the detector and conduct experiments on the M$^3 $FD dataset to evaluate the performance of our EGMT against other SOTA methods in object detection task. Fig.~\ref{fig:object} presents the qualitative comparison results. In the detection of the most prominent pedestrian targets, our EGMT achieves the most complete recognition, while other methods fail to accurately detect overlapping pedestrians. Additionally, in the detection of cars and motorcycles, our EGMT demonstrates competitive detection accuracy. 

Moreover, Table~\ref{object} lists the quantitative comparison results. Most models demonstrate excellent detection performance, with average precision significantly higher than models relying solely on single-modal images. Notably, our EGMT outperforms other competitors, achieving overall average improvements of 1.32${\%}$ and 1.63${\%}$ in mAP@0.5 and mAP@[0.5:0.95], respectively. By effectively leveraging entity-based textual information to enhance visual features, our EGMT significantly improves the overall performance of object detection.

\subsection{Efficiency assessment for image fusion}
We compare the computational efficiency of our EGMT with the other methods on the above four datasets. The comparison covers three key metrics: floating-point operations (FLOPs), parameter count (Params), and inference time on each dataset. To ensure a fair comparison,  we disable the multi-label classification branch of EGMT during all runtime measurements. Results are reported in Table \ref{tab:Efficiency comparison}. Overall, DATFuse attains the lowest computational cost with only 1.185 G FLOPs and 0.011 M parameters, and it is the fastest on TNO and RoadScene. TarDAL achieves the second lowest FLOPs (14.88 G) and parameters (0.296 M), while maintaining competitive inference times on M³FD and MSRS.

In contrast, Text-DiFuse exhibits prohibitive complexity (1,706 G FLOPs, 115.1 M parameters), leading to the longest inference times across all datasets. Although our EGMT involves 111.9 G FLOPs and 53.62 M parameters (higher than most lightweight competitors), its inference latency remains manageable. This cost is deliberate and well-justified, because the cross-modal guided hybrid attention can effectively suppress semantic noise and LVLM hallucinations while injecting entity-level semantics, thereby achieving significant improvements in image quality.

\begin{table*}[ht]
\renewcommand{\arraystretch}{1.3}
\centering
\caption{Efficiency assessment of of our EGMT and the other SOTA methods. 
(MT: multi-task learning; TI: text interaction; 
\textcolor{red}{\textbf{Red}}: optimal; \textcolor{blue}{\textit{Blue}}: sub-optimal)}
\label{tab:Efficiency comparison}

\begin{tabular}{@{}l c c c c c c c c c@{}}
\toprule
\multirow{2}{*}{Methods} &
\multirow{2}{*}{Publishers} &
\multirow{2}{*}{MT} &
\multirow{2}{*}{TI} &
\multirow{2}{*}{FLOPs (G)} &
\multirow{2}{*}{Params. (M)} &
\multicolumn{4}{c}{Inference Time (s)} \\ 
\cmidrule(lr){7-10}
 & & & & & & TNO & RoadScene & M$^3$FD & MSRS \\
\midrule
RFN-Nest \cite{rfn}  & Inf.\ Fus.'21 & \XSolidBrush & \XSolidBrush & 111.1  & 7.524  & 0.143 & 0.127 & 0.163 & 0.136 \\
TarDAL\cite{TAR}     & CVPR'22       & \Checkmark   & \XSolidBrush & \textcolor{blue}{\itshape14.88} & \textcolor{blue}{\itshape0.296} & 0.121 & 0.069 & \textcolor{blue}{\itshape0.148} & \textcolor{blue}{\itshape0.106} \\
SwinFuse\cite{swin}   & TIM'22        & \XSolidBrush & \XSolidBrush & 16.86  & 0.328  & 0.128 & 0.071 & 0.306 & 0.116 \\
DATFuse\cite{DATFuse}    & TCSVT'23      & \XSolidBrush & \XSolidBrush & \textcolor{red}{\bfseries1.185} & \textcolor{red}{\bfseries0.011} & \textcolor{red}{\bfseries0.077} & \textcolor{red}{\bfseries0.043} & 0.153 & 0.088 \\
MetaFusion \cite{meta} & CVPR'23       & \Checkmark   & \XSolidBrush & 53.16  & 0.810  & \textcolor{blue}{\itshape0.118} & 0.064 & \textcolor{red}{\bfseries0.142} & \textcolor{red}{\bfseries0.105} \\
ICAFusion \cite{ICAF} & TMM'23        & \XSolidBrush & \XSolidBrush & 61.31  & 2.471  & 0.136 & 0.076 & 0.300 & 0.125 \\
FILM  \cite{FILM}     & ICML'24       & \XSolidBrush & \Checkmark   & 19.23  & 0.490  & 0.124 & 0.063 & 0.284 & 0.120 \\
Text-DiFuse\cite{Text-diffuse} & NeurIPS'24    & \XSolidBrush & \Checkmark   & 1706   & 115.1  & 690.6 & 7.956 & 707.6 & 14.02 \\
UAAFusion \cite{UAA} & TCSVT'25      & \XSolidBrush & \XSolidBrush & 62.43  & 0.955  & 0.403 & 0.341 & 0.832 & 0.369 \\

Ours       & --             & \Checkmark   & \Checkmark   & 111.9  & 53.62  & 0.181 & 0.071 & 1.483 & 0.161 \\
\bottomrule
\end{tabular}
\end{table*}

\begin{figure*}[ht]
  \centering
  \includegraphics[width=\linewidth]{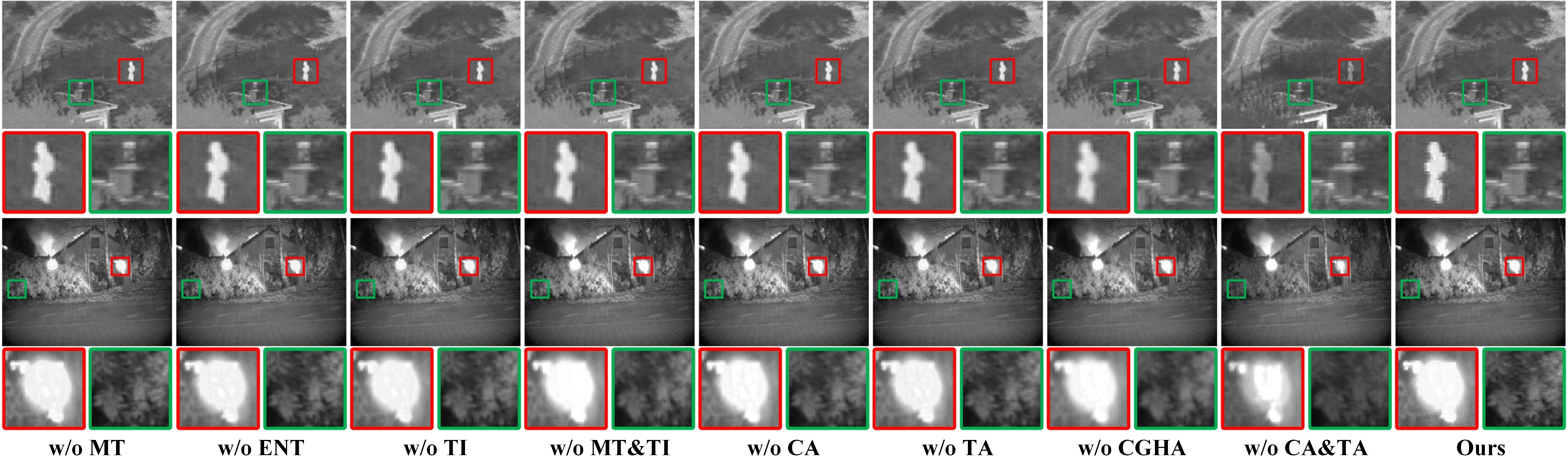}
\caption{Qualitative assessment of our EGMT and its variants on the TNO dataset.}
  \label{figablation}
\end{figure*}

\subsection{Ablation studies}

We conduct ablation studies on the TNO dataset to validate the contributions of two core components: the entity-guided multi-task learning framework and the entity-guided interaction module (ECIM). By systematically removing specific designs, we quantitatively and qualitatively analyze their impacts on fusion performance.

\subsubsection{Analysis of entity-guided multi-task learning} 

In the proposed EGMT, we extract entity-level textual information and use entities as pseudo-labels for multi-task learning. This not only enhances the model's understanding of key visual features but also significantly improves the semantic density of the fused images. 

The qualitative assessment is shown in Fig.\ref{figablation}, where the full EGMT model retains better target profile and brightness distribution compared to these degraded variants. The quantitative assessment is shown in Table \ref{tab:ablation}, removing the multi-task learning framework (w/o MT) leads to significant performance drops, with PC decreasing by 12.1\%, MI by 16.5\%, and Q\(_{abf}\) by 7.1\%. This highlights that the auxiliary multi-label classification task enhances semantic perception by leveraging entity pseudo-labels. When entity extraction is disabled (w/o ENT), the model relies on raw LVLM-generated captions, resulting in increased structural noise (N\(_{abf}\) rises from 0.00595 to 0.01476). This indicates that raw text contains redundant linguistic noise, which interferes with precise feature interaction. Removing text interaction (w/o TI) causes a marked decline in FMI\(_w\) (26.8\%) and MI (23.1\%), underscoring the importance of text-driven semantic guidance. Furthermore, removing both MT and TI (w/o MT\&TI) leads to catastrophic performance drops across all metrics, confirming that semantic supervision from entity-guided fusion and classification tasks is critical for high-quality fused results.

\begin{table*}[!t]
	\renewcommand\arraystretch{1.5}
        \caption{Quantitative assessment of our EGMT and its variants on the TNO dataset. (MT: multi-task learning; ENT: entity; TI: text interaction; CA: channel-wise attention, TA: token-wise attention; CGHA: cross-modal guided hybrid attention; \textcolor{red}{\textbf{Red}}: optimal; \textcolor{blue}{\textit{Blue}}: sub-optimal)}
	\centering 
	\begin{tabular}{ l c c c c c c c c c c c c c }
		\hline
        Models & CA & TA & CGHA & MT & TI & PC ${\uparrow}$ & SSIM ${\uparrow}$ & MI ${\uparrow}$ & Q$_{abf}$ ${\uparrow}$  & PSNR ${\uparrow}$ & FMI$_{w}$ ${\uparrow}$ & N$_{abf}$ ${\downarrow}$  & NCIE ${\uparrow}$   \\
		\hline
		w/o MT & \Checkmark & \Checkmark & \Checkmark  & \XSolidBrush & \Checkmark  & \textcolor{blue}{\itshape0.35365} & \textcolor{red}{\bfseries0.72371} & \textcolor{blue}{\itshape4.06336} & \textcolor{blue}{\itshape0.52417} & \textcolor{red}{\bfseries16.53178} & 0.38242 &\textcolor{blue}{\itshape 0.01020} & \textcolor{blue}{\itshape0.81206}	\\
        
        w/o ENT & \Checkmark & \Checkmark & \Checkmark &  \XSolidBrush & \XSolidBrush & 0.34562 & 0.71912 & 4.02937 & 0.52031 & 16.43338 & 0.40343 & 0.01476 & 0.81200        \\
		w/o TI & \Checkmark & \Checkmark & \Checkmark  &  \Checkmark & \XSolidBrush & 0.31214 & 0.71947 & 3.73969 & 0.47894 & \textcolor{blue}{\itshape16.46733} & 0.33653 & 0.01039 & 0.81032  \\

        w/o MT\&TI & \Checkmark & \Checkmark & \Checkmark & \XSolidBrush & \XSolidBrush & 0.30418 & 0.70862 & 3.79731 & 0.48545 & 14.14679 & 0.39675 & 0.02462 & 0.81108       \\

        \hline
        
		w/o CA & \XSolidBrush  & \Checkmark & \Checkmark & \Checkmark & \Checkmark  & 0.31459 & 0.72005 & 3.73679 & 0.50434 & 16.28639 & \textcolor{blue}{\itshape0.40728} & 0.01646 & 0.81064 \\
        
        w/o TA & \Checkmark & \XSolidBrush  & \Checkmark  & \Checkmark & \Checkmark & 0.31189 & 0.72071 & 3.63799 & 0.48363 & 16.40115 & 0.31814 & 0.01384 & 0.80986		\\

	   w/o CGHA & \Checkmark & \Checkmark & \XSolidBrush  & \Checkmark & \Checkmark & 0.27508 & 0.71469 & 3.50758 & 0.46008 & 16.23785 & 0.34428 & 0.01487 & 0.80960	 \\
        
        w/o CA\&TA & \XSolidBrush  & \XSolidBrush  & \Checkmark  & \Checkmark & \Checkmark & 0.33971 & 0.70971 & 2.65552 & 0.48184 & 16.24159 & 0.37570 & 0.01236 & 0.80668		\\

		Ours & \Checkmark & \Checkmark & \Checkmark & \Checkmark & \Checkmark & \textcolor{red}{\bfseries0.40241} & \textcolor{blue}{\itshape0.72085} & \textcolor{red}{\bfseries4.86354} & \textcolor{red}{\bfseries0.56391} & 16.46613 & \textcolor{red}{\bfseries0.45979} & \textcolor{red}{\bfseries0.00595} & \textcolor{red}{\bfseries0.81674} \\
		\hline 		
        \label{tab:ablation}
	\end{tabular}
\end{table*}

\subsubsection{Analysis of ECIM} 

The ECIM employs hierarchical attention mechanisms to boost the fine-grained cross-modal feature interaction between inter image and visual-entity levels. This enables the model to concentrate on relevant visual regions guided by entities, ensuring a thorough comprehension of both visual content and contextual information.

As illustrated in Fig.\ref{figablation}, the quality of the fused results of different variants is impacted by varied levels of degradation. Specifically, the variant without CGHA lacks the ability to regulate the apparent intensity, while the variant without CA\&TA generates unbalanced outcomes. As shown in Table \ref{tab:ablation}, removing channel-wise attention (w/o CA) causes a 23.2\% drop in MI, indicating damaged cross-modal information transmission between infrared and visible modalities due to lost channel-level dependencies. Disabling token-wise attention (w/o TA) mainly impacts edge preservation, as shown by a 14.2\% drop in Q\(_{abf}\), leading to fragmented feature maps and blurred boundaries in the fused results. Removing cross-modal guided hybrid attention (CGHA) results in the most balanced performance degradation (PC↓31.6\%, MI↓27.9\%), indicating its core role in the visual-entity semantic interaction: by alternating guidance, it suppresses irrelevant entity interference with the image content. Jointly removing CA and TA leads to a 45.4\% MI reduction, highlighting their collective importance for comprehensive cross-modal interaction. Variants also show varying SSIM and PSNR drops, confirming cross-modal interactions' contribution to structural consistency, detail fidelity, and overall quality optimization. This reinforces the ECIM's irreplaceability in coordinating semantic guidance and low-level feature enhancement.

\subsection{Limitations}~\label{sec:limitation}

While EGMT demonstrates the SOTA performance across multiple benchmarks, we acknowledge two limitations that warrant discussion. Firstly, the current framework relies on a pre-trained LVLM and entity detector for text generation and entity extraction, which introduces computational overhead during preprocessing. Although this dependency between visual and entity features ensures high-quality semantic guidance, it may limit the deployment in resource-constrained environments. Besides, the multi-task learning framework also increases the complexity of the network. Secondly, while EGMT excels in general scenarios, its performance in extreme conditions (\textit{e.g.}, severe motion blur and ultra-low-light environments) requires further validation. However, our approach's entity-guided design inherently prioritizes semantically salient regions, demonstrating potential advantages in such challenging cases that merit dedicated investigation. 

Although these limitations still exist, EGMT represents a significant advancement by successfully bridging the semantic gap between textual entities and visual fusion process. The consistent performance gains across diverse datasets and downstream task validate our core hypothesis that entity-guided multi-task learning enables more semantically coherent fusion. The framework's modular architecture permits straightforward integration of improved entity detectors or large vision-language models as they emerge, ensuring enduring relevance in this rapidly evolving field.

\section{Conclusion}~\label{sec:conclusion}

In this paper, we propose EGMT, an entity-guided multi-task learning framework for infrared and visible image fusion. Our key innovation involves leveraging entity-level semantics to address text integration challenges in fusion tasks. We establish a principled entity extraction method that eliminates semantic noise while preserving critical information, and develop a multi-task architecture that enhances semantic perception through auxiliary classification using entity pseudo-labels. The proposed cross-modal interaction module enables hierarchical feature fusion at both inter-visual and visual-entity levels. 
Our experimental findings demonstrate that entity-level guidance provides a powerful supervisory signal, offering more interpretable and controllable fusion compared to sentence-level or task-specific semantic approaches. Extensive evaluations show that EGMT outperforms state-of-the-art methods in fusion quality while maintaining strong classification performance, establishing a new paradigm for semantic-guided image fusion.

Future work will focus on developing lightweight architectures and enhancing robustness under extreme conditions to improve practical applicability.

{
    \small
    \bibliographystyle{IEEEtran}
    \bibliography{main}
}

\vfill

\end{document}